\documentclass[square,comma]{article}

\usepackage[square, comma, sort&compress, numbers]{natbib}
\usepackage[final]{bdl_2018}

\usepackage[utf8]{inputenc} % allow utf-8 input
\usepackage[T1]{fontenc}    % use 8-bit T1 fonts
\usepackage{hyperref}       % hyperlinks
\usepackage{url}            % simple URL typesetting
\usepackage{booktabs}       % professional-quality tables
\usepackage{amsfonts}       % blackboard math symbols
\usepackage{nicefrac}       % compact symbols for 1/2, etc.
\usepackage{microtype}      % microtypography
\usepackage{amssymb}
\usepackage{amsmath}
\usepackage{bm}
\usepackage{subfig}
\usepackage{wrapfig}
\usepackage{lipsum}     
\usepackage{float} 
\usepackage{graphicx}           
\usepackage{url} 
\usepackage{soul}
\usepackage{enumitem}
\usepackage{lineno}
\usepackage{color}
\usepackage[title]{appendix}
\usepackage{diagbox}

\title{Physics-informed deep generative models}

\author{
Yibo Yang, Paris Perdikaris\\
Department of Mechanical Engineering and Applied Mechanics, \\
University of Pennsylvania, \\
Philadelphia, PA 19104, USA \\
\texttt{\{ybyang, pgp\}@seas.upenn.edu} \\
}

\begin{document}

\maketitle

\begin{abstract}
We consider the application of deep generative models in propagating  uncertainty through complex physical systems. Specifically, we put forth an implicit variational inference formulation that constrains the generative model output to satisfy given physical laws expressed by partial differential equations. Such physics-informed constraints provide a regularization mechanism for effectively training deep probabilistic models for modeling physical systems in which the cost of data acquisition is high and training data-sets are typically small. This  provides a scalable framework for characterizing  uncertainty in the outputs of physical systems due to randomness in their inputs or noise in their observations. We demonstrate the effectiveness of our approach through a canonical example in transport dynamics.
\end{abstract}

\section{Introduction}
Despite their immense recent success, many state-of-the art machine learning techniques (e.g., deep neural nets, convolutional networks, recurrent networks \citep{1,2,3,4}, etc.) are lacking robustness and often fail to provide any guarantees of convergence or quantify the error/uncertainty associated with their predictions. Hence, the ability to quantify predictive uncertainty and learn in a sample-efficient manner is a necessity \citep{5,6,7,8}, especially in data-limited domains \citep{9}. Even less well understood is how one can constrain such algorithms to leverage domain-specific knowledge and return predictions that satisfy certain physical principles \citep{10} (e.g., conservation of mass, momentum, etc.). In recent work, Raissi {\em et. al.}  \citep{11,12,13,14} explored this new interface between classical scientific computing and machine learning by revisiting the idea of penalizing the loss function of deep neural networks using differential equation constraints, as first put forth by Psichogios and Ungar \citep{15} and Lagaris {\em et. al.} \citep{16}. In this work, we revisit these approaches from a probabilistic standpoint \citep{17,18,19} and develop an adversarial inference framework \citep{20,21,22,23} to enable the  posterior characterization \citep{24,25} of the uncertainty associated with the model predictions. These uncertainty estimates reflect how observation noise and/or randomness in the system's inputs and outputs are propagated through complex infinite-dimensional dynamical systems described by non-linear partial differential equations (PDEs).

\section{Methods}
\label{sec:Methods}
In Raissi {\em et. al.} \citep{11,12,13,14}, the authors have considered constructing deep neural networks that return predictions which are constrained by PDEs of the form
$\bm{u}_t + \mathcal{N}_{\bm{x}}\bm{u} = 0$, 
where $\bm{u}(\bm{x},t)$ is represented by a deep neural network parametrized by a set of parameters $\theta$, i.e. $\bm{u}(\bm{x},t) = f_{\theta}(\bm{x},t)$, $\bm{x}$ is a vector of space coordinates, $t$ is the time coordinate, and $\mathcal{N}_{\bm{x}}$ is a nonlinear differential operator. As neural networks are differentiable representations, this construction defines a so-called {\em physics informed neural network} that corresponds to the PDE residual, i.e. $\bm{r}_{\theta}(\bm{x},t):= \frac{\partial}{\partial t} f_{\theta}(\bm{x},t)  + \mathcal{N}_{\bm{x}}f_{\theta}(\bm{x},t)$.
By construction, this network shares the same architecture and parameters with $f_{\theta}(\bm{x},t)$, but it has different activation functions corresponding to the action of the differential operator. 
The resulting training procedure allows us to recover the shared network parameters $\theta$ using a few scattered observations of $\bm{u}(\bm{x},t)$, namely $\{(\bm{x}_{i}, t_i), \bm{u}_i\}$, $i = 1,\dots,N_u$, along with a larger number of collocation points $\{(\bm{x}_{i}, t_i), \bm{r}_i = 0\}$, $i = 1,\dots,N_r$, that aim to penalize the PDE residual at a finite set of $N_r$ collocation nodes. This way, the resulting optimization problem can be effectively solved using standard stochastic gradient descent without necessitating any elaborate constrained optimization techniques, simply by minimizing the composite loss function
\begin{equation}\label{eq:PINN_loss}
\mathcal{L}_{\text{PDE}}(\theta) := \frac{1}{N_u}\sum\limits_{i=1}^{N_u}\|f_{\theta}(\bm{x}_{i}, t_i) - \bm{u}_i\|^2  +   \frac{1}{N_r}\sum\limits_{i=1}^{N_r}\|\bm{r}_{\theta}(\bm{x}_{i}, t_i) - \bm{r}_i\|^2, 
\end{equation}
where the required gradients can be readily obtained using automatic differentiation \citep{26}.

In the proposed work we aim to model uncertainty in such physics-informed models by constructing conditional latent variable  models of the form
\begin{equation}
\label{eq:PSC}
p(\bm{u}|\bm{x},t) = \int p(\bm{u},\bm{z}|\bm{x},t) d\bm{z} = \int p(\bm{u}|\bm{x},t,\bm{z}) p(\bm{z}|\bm{x},t) d\bm{z},
\end{equation}
and encourage the resulting samples to be constrained by a given physical law according to a likelihood
\begin{equation}
\label{eq:PIDGM}
p(\bm{u}|\bm{x},t,\bm{z}), \ \ \bm{z}\sim p(\bm{z}), \ \ \text{such that} \ \  \bm{u}_t + \mathcal{N}_{\bm{x}}\bm{u} = 0.
\end{equation}
This setting encapsulates a wide range of deterministic and stochastic problems, where $\bm{u}(\bm{x},t)$ is a potentially multi-variate field, and $\bm{z}$ is a collection of latent variables.

Following the recent findings of \citep{27} we will train the generative model by matching the joint distribution of the generated samples  $p_{\theta}(\bm{x},t,\bm{u})$ with the joint distribution of the observed data $q(\bm{x},t,\bm{u})$ by 
 minimizing the reverse Kullback-Leibler (KL) divergence, which can be decomposed as
\begin{align}\label{eq:KL}
\mathbb{KL}[p_{\theta}(\bm{x},t,\bm{u})||q(\bm{x},t,\bm{u})] & = -h(p_{\theta}(\bm{x},t,\bm{u}))) - \mathbb{E}_{p_{\theta}(\bm{x},t,\bm{u})}[\log(q(\bm{x},t,\bm{u}))],
\end{align}
where $h(p_{\theta}(\bm{x},t,\bm{u}))$ denotes the entropy of the generative model. This decomposition reveals the interplay between two competing mechanisms that define the model training dynamics. On one hand, minimizing the negative entropy term is encouraging the support of $p_{\theta}(\bm{x},t,\bm{u})$ to spread to infinity, while the second term will penalize regions for which the support of $p_{\theta}(\bm{x},t,\bm{u})$ and $q(\bm{x},t,\bm{u})$ do not overlap. As observed in \citep{27}, this introduces a regularization mechanism for mitigating the pathology of mode collapse.

As the entropy term $h(p_{\theta}(\bm{x},t,\bm{u})))$ is intractable, we can obtain a computable training objective by deriving the following lower bound (see proof in Appendix A)
\begin{equation}
    h(p_{\theta}(\bm{x},t,\bm{u})) \ge h(p(\bm{z})) + \mathbb{E}_{p_{\theta}(\bm{x}, t, \bm{u}, \bm{z})}[\log(q_{\phi}(\bm{z}|\bm{x}, t, \bm{u}))],
\end{equation}
where the inference model $q_{\phi}(\bm{z}|\bm{x}, t, \bm{u})$ plays the role of a variational approximation to the true posterior over the latent variables, and appears naturally using information theoretic arguments in the derivation of the lower bound \citep{27}.

This construction leads to two coupled objectives that define an  adversarial game for training the model parameters
\begin{align}
	\mathcal{L}_{\mathcal{D}}(\psi) := & \  \mathbb{E}_{q(\bm{x},t)p(\bm{z})}[\log\sigma(T_{\psi}(\bm{x},t,f_{\theta}(\bm{x},t,\bm{z})))] + \nonumber \\ & \ \mathbb{E}_{q(\bm{x},t,\bm{u})}[\log(1-\sigma(T_{\psi}(\bm{x},t,\bm{u})))] \label{eq:discriminator_loss}\\
	\mathcal{L}_{\mathcal{G}}(\theta, \phi) := & \ \mathbb{E}_{q(\bm{x},t)p(\bm{z})}[T_{\psi}(\bm{x}, t, f_{\theta}(\bm{x},t,\bm{z}))+ (1-\lambda)\log(q_{\phi}(\bm{z}|\bm{x},t,f_{\theta}(\bm{x},t,\bm{z})))] \label{eq:generator_loss},
\end{align}
where $T_{\psi}(\bm{x},t,\bm{u})$ is a parametrized discriminator used to approximate the KL divergence directly from samples using the density ratio trick \citep{28}, and $\sigma(x)=1/(1+e^{-x})$ is the logistic sigmoid function. Moreover, notice how the inference model  $q_{\phi}(\bm{z}|\bm{x},t,f_{\theta}(\bm{x},t,\bm{z}))$ promotes cycle consistency in the latent variables, and serves as an entropic regularization term  than allows us to stabilize model training and mitigate the pathology of mode collapse, as controlled by the user defined parameter $\lambda$ \citep{27}. 
The final training objective that encourages the generated samples to satisfy a given PDE reads as
\begin{equation}
\label{eq:advi_objective}
    \begin{aligned}
    	& \mathop{\max}_{\psi} \ \mathcal{L}_{\mathcal{D}}(\psi)\\
    	& \mathop{\min}_{\theta, \phi} \ \mathcal{L}_{\mathcal{G}}(\theta, \phi) + \beta \mathcal{L}_{\text{PDE}}(\theta),
    \end{aligned}
\end{equation}
where positive values of $\beta$ can be selected to place more emphasis on penalizing the PDE residual. For $\beta>0$, the residual loss $\mathcal{L}_{\text{PDE}}(\theta)$ acts as a regularization term that encourages the generator $p_{\theta}(\bm{u}|\bm{x},t,\bm{z})$ to produce samples that satisfy the underlying partial differential equation.

\section{Results}
\label{sec4}
Here we demonstrate the performance of the proposed methodology through the lens of a canonical problem in transport dynamics modeled by the Burgers equation with appropriate initial and boundary conditions. This equation arises in various areas of applied mathematics, including fluid mechanics, nonlinear acoustics, gas dynamics, and traffic flow \citep{29}. In one space dimension the equation reads as

\begin{equation}
\label{eq:Burgers}
\begin{aligned}
	&u_t + u u_x - \nu u_{xx} = 0, \quad x\in[-1, 1], \quad  t\in[0,1],\\
	&u(0,x) = -\sin(\pi x), \quad u(t,-1) = u(t,1) = 0,\\
\end{aligned}
\end{equation}
where $\nu = 0.01/\pi$ is a viscosity parameter, small values of which can lead to solutions developing shock formations that
are notoriously hard to resolve by classical numerical methods \citep{29}.

Here, we construct a probabilistic representation for the unknown solution $u(x,t)$ using a physics-informed deep generative model $p_{\theta}(u|x,t,z)$, and we introduce parametric mappings corresponding to a generator $f_{\theta}(x, t, z)$, an encoder $q_{\phi}(z|x, t, u)$, and a discriminator $T_{\psi}(x, t, u)$ all constructed using deep feed-forward neural networks with 4 hidden layers with 50 neurons each. The activation function in all cases is chosen to be a hyperbolic tangent non-linearity. The prior over the latent variables is chosen  to be a one-dimensional isotropic Gaussian distribution, i.e. $z \sim \mathcal{N}(0,1)$. We train our probabilistic model using $30,000$ stochastic gradient Adam updates \citep{30} using a learning rate of $10^{-4}$ on a data-set comprising of $N_u = 200$ input/output pairs for $u(x,t)$ -- 100 points for the initial condition and 50 points for each of the domain boundaries -- plus an additional  $N_r = 10,000$ collocation points for enforcing the residual of the Burgers equation using the loss of equation \ref{eq:advi_objective} with $\lambda = 1.5$ and $\beta = 1.0$. A systematic study with respect to the model hyper-parameters is provided in Appendix B. Notice that the initial condition here is corrupted by a non-additive noise process as
\[
u(x, 0) = -\sin(\pi(x+2\delta))+\delta, \qquad \delta = \epsilon/\exp(3|x|), \qquad \epsilon\sim \mathcal{N}(0, 0.1^2),
\]
and our goal here is to propagate the effect of this uncertainty into the prediction of future system states. Notice how the noise variance is chosen to be larger around $x=0$, therefore amplifying the effect of uncertainty on the shock formation.

\begin{figure}[b]
\centering
\includegraphics[width=\textwidth]{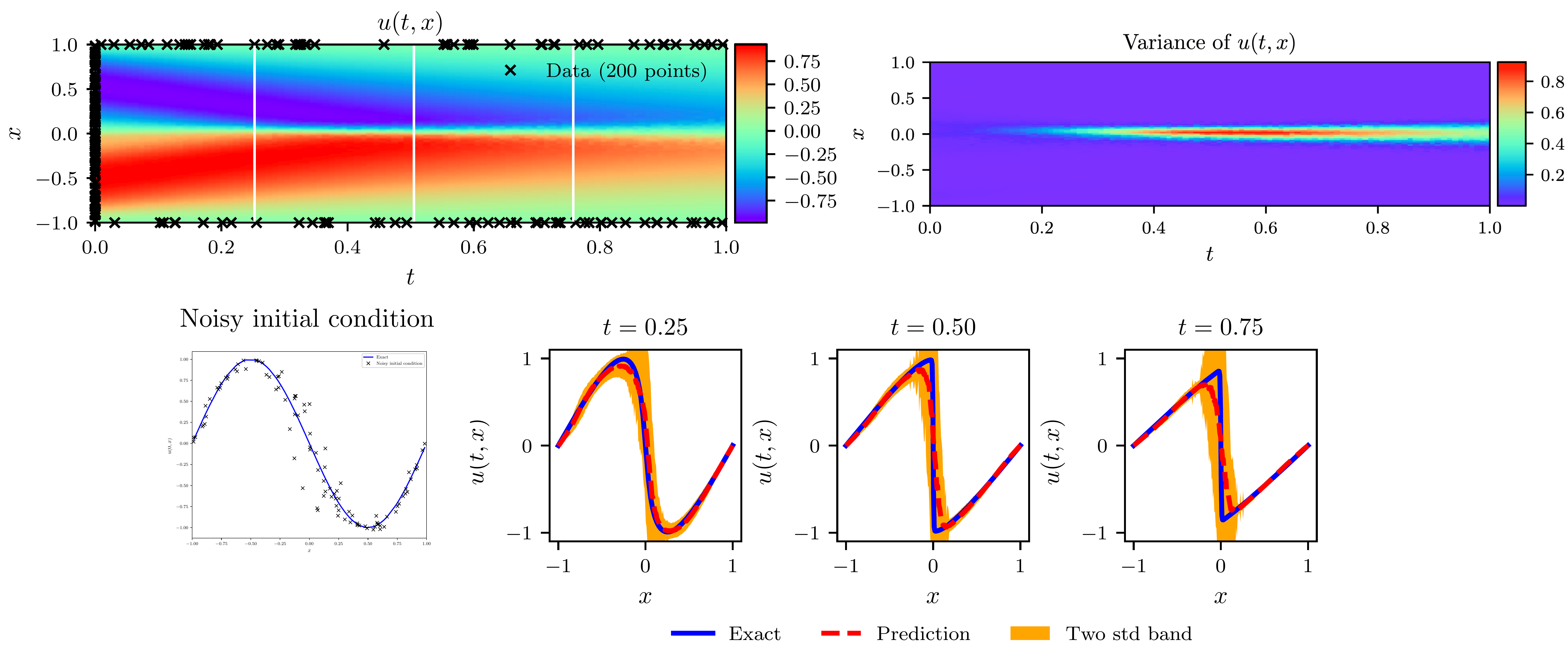}
\caption{{\it Top:} Mean and variance of $p_{\theta}(u|x,t,z)$, along with the location of the noisy training data $\{(x_{i}, t_i), u_i\}_{i=1}^{N_u}$. {\it Bottom:} Noisy data for the initial condition, and the resulting prediction and predictive uncertainty at $t=0.25$, $t=0.5$ and $t=0.75$.}
\label{fig:burgers_noisy}
\end{figure}

The results of this experiment are summarized in figure  \ref{fig:burgers_noisy}, where we report the predicted mean solution, as well as the uncertainty associated with this prediction. We observe that the resulting generative model $p_{\theta}(u|x,t,z)$ can effectively capture the uncertainty in the resulting spatio-temporal solution due to the propagation of the input noise process through the complex non-linear dynamics of the Burgers equation. As expected, the uncertainty concentrates around the shock discontinuity that the solution develops around $t=0.5$. Although we only plot the first two moments of the solution, we must emphasize that the generative model $p_{\theta}(u|x,t,z)$ provides a complete probabilistic characterization of its non-Gaussian statistics.

% \vspace{-0.7cm}
%The results shows that we get accurate prediction on the solution of Burgers equations. Especially, the uncertainty we derive capture the shock of the wave precisely. 

\clearpage

%\section*{References}

%\begin{appendices}
\clearpage

\appendix

\section{Proof of the Entropy Lower Bound}
Here we follow the derivation of Li \citep{27} to construct a computable lower bound for the entropy $h(p_{\theta}(\bm{x},t,\bm{u})))$. To this end, we start by considering random variables $(\bm{x}, t, \bm{u}, \bm{z})$ under the joint distribution 
$$p_{\theta}(\bm{x}, t, \bm{u}, \bm{z}) = p_{\theta}(\bm{u},\bm{x}, t| \bm{z}) p(\bm{z}) = p_{\theta}(\bm{u}|\bm{x}, t, \bm{z})p(\bm{x}, t) p(\bm{z}),$$ 
where $p_{\theta}(\bm{u}|\bm{x}, t, \bm{z}) = \delta(\bm{u}-f_{\theta}(\bm{x}, t, \bm{z}))$, and $\delta(\cdot)$ is the Dirac delta function. The mutual information between $(\bm{x}, t, \bm{u})$ and $\bm{z}$ satisfies the information theoretic identity
$$ I(\bm{x}, t, \bm{u}; \bm{z}) = h(\bm{x}, t, \bm{u})-h(\bm{x}, t, \bm{u}|\bm{z}) =
h(\bm{z}) - h(\bm{z}|\bm{x}, t, \bm{u}),$$
where $h(\bm{x}, t, \bm{u})$, $h(\bm{z})$ are the marginal entropies and $h(\bm{x}, t, \bm{u}|\bm{z})$, $h(\bm{z}|\bm{x}, t, \bm{u})$ are the conditional entropies \citep{31}. 
Since in our setup $\bm{x}$ and $t$ are deterministic variables independent of $\bm{z}$, and samples of $p_{\theta}(\bm{u}|\bm{x}, t, \bm{z})$ are generated by a deterministic function $f_{\theta}(\bm{x}, t, \bm{z})$, it follows that $h(\bm{x}, t, \bm{u}|\bm{z}) = 0$. 
We therefore have
\begin{equation}\label{eq:info_identity}
h(\bm{x}, t, \bm{u}) = h(\bm{z}) - h(\bm{z}|\bm{x}, t, \bm{u}),
\end{equation}
where $h(z) := -\int \log p(\bm{z}) p(\bm{z}) d\bm{z}$ is a constant with respect to the model parameters $\theta$. 

Now consider a general variational distribution $q_{\phi}(\bm{z}|\bm{x}, t, \bm{u})$ parametrized by a set of parameters $\phi$. Then,

\begin{align}
   h(\bm{z}|\bm{x}, t, \bm{u})  = & -\mathbb{E}_{p_{\theta}(\bm{x}, t, \bm{u}, \bm{z})}[\log(p_{\theta}(\bm{z}|\bm{x}, t, \bm{u}))] \nonumber\\
    = & -\mathbb{E}_{p_{\theta}(\bm{x}, t, \bm{u}, \bm{z})}[\log(q_{\phi}(\bm{z}|\bm{x}, t, \bm{u}))] \nonumber  \\ 
    & -\mathbb{E}_{p_{\theta}(\bm{x}, t, \bm{u})}[\mathbb{KL}[p_{\theta}(\bm{z}|\bm{x}, t, \bm{u})||q_{\phi}(\bm{z}|\bm{x}, t, \bm{u})]] \nonumber \\
    \le & -\mathbb{E}_{p_{\theta}(\bm{x}, t, \bm{u}, \bm{z})}[\log(q_{\phi}(\bm{z}|\bm{x}, t, \bm{u}))]. \label{eq:entropy_bound}
\end{align}
Viewing $\bm{z}$ as a set of latent variables, then $q_{\phi}(\bm{z}|\bm{x}, t, \bm{u})$ is a variational approximation to the true intractable posterior over the latent variables $p_{\theta}(\bm{z}|\bm{x}, t, \bm{u})$. Therefore, 
if $q_{\phi}(\bm{z}|\bm{x}, t, \bm{u})$  is introduced as an auxiliary encoder associated with the generative model $p_{\theta}(\bm{x}, t, \bm{u})$, for which $\bm{u} = f_{\theta}(\bm{x}, t,\bm{z})$ and $\bm{z}\sim p(\bm{z})$, then we can use equations \ref{eq:info_identity} and \ref{eq:entropy_bound} to bound the entropy term in equation \ref{eq:KL} as
\begin{equation}
    h(p_{\theta}(\bm{x},t,\bm{u})) \ge h(p(\bm{z})) + \mathbb{E}_{p_{\theta}(\bm{x}, t, \bm{u}, \bm{z})}[\log(q_{\phi}(\bm{z}|\bm{x}, t, \bm{u}))]
\end{equation}

\section{Systematic Studies}

Here we provide results on a series of comprehensive systematic studies that aim to quantify the sensitivity of the resulting predictions on: (i) the neural network initialization, (ii) the total number of training and collocation points, (iii) the neural network architecture, and (iv) the adversarial training procedure. In all cases we have used the non-linear Burgers equation as a prototype problem.

\subsection{Sensitivity with respect to the neural network initialization}

In order to quantify the sensitivity of the proposed methods with respect to the initialization of the neural networks, we have considered a noise-free data set comprising of $N_u = 150$ and $N_r = 10000$ training and collocation points, respectively, and fixed the architecture for generator neural networks to include 4 hidden layers with 50 neurons each and discriminator neural networks to include 3 hidden layers with 50 neurons each
%all neural networks to include 4 hidden layers with 50 neurons each%
, and a hyperbolic tangent activation function. Then we have trained an ensemble of 15 cases all starting from a normal Xavier initialization \citep{32} for all network weights (with a randomized seed), and a zero initialization for all bias parameters. In table \ref{tab:sens_t1} we report the relative error between the predicted mean solution and the known exact solution for this problem for all 15 randomized trials using at set of $25600$ randomly selected test points. Evidently, our results are robust with respect to the the neural network initialization as in all cases the stochastic gradient descent training procedure converged roughly to the same solution. We can summarize this result by reporting the mean and the standard deviation of the relative $\mathcal{L}_2$ error as 
\[
\hat{\mathcal{L}_2} \in [\mu_L-\sigma_L, \mu_L+\sigma_L] = [4.7\times 10^{-2} - 1.3\times 10^{-2}, 4.7\times 10^{-2} + 1.3\times 10^{-2}].
\]

\begin{table}[!htbp]
\centering
\begin{tabular}{|c|c|c|c|c|c|}
\hline
\multicolumn{5}{|c|}{Relative $\mathcal{L}_2$ error}\\ 
\hline
4.1e-02& 7.9e-02& 4.4e-02& 4.0e-02& 3.8e-02\\
\hline
3.2e-02& 5.7e-02& 4.7e-02& 6.5e-02& 4.0e-02\\
\hline
3.5e-02& 3.5e-02& 6.4e-02& 4.0e-02& 4.9e-02\\
\hline
\end{tabular}
\caption{Relative $\mathcal{L}_2$ error with different seed of initialization in non-noise case.}
\label{tab:sens_t1}
\end{table}

\subsection{Sensitivity with respect to the total number of training and collocation points}\label{B2}

In this study our goal is to quantify the sensitivity of our predictions with respect to the total number of training and collocation points $N_u$ and $N_r$, respectively. As before, we have considered noise-free data sets, and fixed the architecture for generator neural networks to include 4 hidden layers with 50 neurons each and discriminator neural networks to include 3 hidden layers with 50 neurons each, a hyperbolic tangent activation function, and a normal Xavier initialization \citep{32} for all network weights and zero initialization for all network biases. The results of this study are summarized in table \ref{tab:sens_t2}, indicating that as the number of collocation points are increased, a more accurate prediction is obtained. This observation is in agreement with the original results of Raissi {\em et. al.} \citep{11} for deterministic physics-informed neural networks, indicating the role of the PDE residual loss as an effective regularization mechanism for training deep generative models in small data regimes.

\begin{table}[!htbp]
\centering
\begin{tabular}{|c|c|c|c|c|c|c|c|c|c|}
\hline
\diagbox{$N_u$}{$N_r$} & 10& 100& 250& 500& 1000& 5000& 10000\\ 
\hline
60  & 9.3e-01& 5.6e-01& 4.8e-01& 5.0e-02& 1.9e-01& 5.0e-02& 5.1e-02\\
\hline
90  & 5.8e-01& 5.3e-01& 3.5e-01& 1.5e-01& 4.9e-02& 1.0e-01& 5.8e-02\\
\hline
150 & 6.7e-01& 1.4e-01& 3.0e-01& 3.6e-02& 4.9e-02& 1.2e-01& 4.7e-02\\
\hline
\end{tabular}
\caption{Relative $\mathcal{L}_2$ prediction error for different number of training and collocation points $N_u$ and $N_r$, respectively.}
\label{tab:sens_t2}
\end{table}

\subsection{Sensitivity with respect to the neural network architecture}

In this study we aim to quantify the sensitivity of our predictions with respect to the architecture of the neural networks that parametrize the generator, the discriminator, and the encoder. Here we have fixed the number of noise-free training data to $N_u = 150$ and $N_r =10000$, and we kept the number of layers for discriminator to always be one less than the number of layers for generator (e.g., if the number of layers for generator is two then the number of layers for discriminator is one, etc.). 
% {\color{red}{This choice is related to the discussion provided in section ?? related to the stability of the  adversarial training process.)}}
% {\color{blue}{\text{We do not have this context in the article, how about delete this line?)}}} 
In all cases, we have used a hyperbolic tangent non-linearity and a normal Xavier initialization \citep{32}. In table \ref{tab:sens_t3} we report the relative $\mathcal{L}_2$ prediction error for different feed-forward architectures for the generator, discriminator, and encoder (i.e., different number of layers and number of nodes in each layer). The general trend suggests that as the neural network capacity is increased we obtain more accurate predictions, indicating that our physics-informed constraint on the PDE residual can effectively regularize the training process and safe-guard against over-fitting. We note number of neurons in each layer as $N_{n}$ and number of layers for generator (encoder) as $N_g$.

\begin{table}[!htbp]
\centering
\begin{tabular}{|c|c|c|c|}
\hline
\diagbox{$N_g$}{$N_n$} & 20& 50& 100\\ 
\hline
2 &    4.2e-01& 3.8e-01& 5.7e-01 \\
\hline
3 &    6.5e-02& 3.5e-02& 2.1e-02 \\
\hline
4 &    9.3e-02& 4.7e-02& 5.4e-02 \\
\hline
\end{tabular}
\caption{Relative $\mathcal{L}_2$ prediction error for different feed-forward architectures for the generator, encoder, and the discriminator. The total number of layers of the latter was always chosen to be one less than the number of layers for generator.}
\label{tab:sens_t3}
\end{table}

\subsection{Sensitivity with respect to the adversarial training procedure}

Finally, we test the sensitivity with respect to the adversarial training process. To this end, we have fixed the number of noise-free training data to $N_u = 150$ and $N_r =10000$, and the neural network architecture to be the same as \ref{B2}, and we vary the total number of training steps for the generator $K_g$ and the discriminator $K_d$ within each stochastic gradient descent iteration. The results of this study are presented in table \ref{tab:sens_t4} where we report the relative $\mathcal{L}_2$ prediction error. These results reveal the high sensitivity of the training dynamics on the interplay between the generator and discriminator networks, and pinpoint on the well known peculiarity of adversarial inference procedures which require a careful tuning of 
$K_g$ and $K_d$ for achieving stable performance in practice.

\begin{table}[!htbp]
\centering
\begin{tabular}{|c|c|c|c|}
\hline
\diagbox{$K_g$}{$K_d$} & 1& 2& 5\\ 
\hline
1 &    3.5e-01& 5.0e-01& 1.5e+00 \\
\hline
2 &    4.3e-02& 3.2e-01& 5.4e-01 \\
\hline
5 &    4.7e-02& 2.3e-01& 7.0e-01 \\
\hline
\end{tabular}
\caption{Relative $\mathcal{L}_2$ error with different number of training for generator and discriminator in each epoch.}
\label{tab:sens_t4}
\end{table}

%\end{appendices}

%[1] Alexander, J.A.\ \& Mozer, M.C.\ (1995) Template-based algorithms
%for connectionist rule extraction. In G.\ Tesauro, D.S.\ Touretzky and
%T.K.\ Leen (eds.), {\it Advances in Neural Information Processing
%  Systems 7}, pp.\ 609--616. Cambridge, MA: MIT Press.

%[2] Bower, J.M.\ \& Beeman, D.\ (1995) {\it The Book of GENESIS:
%  Exploring Realistic Neural Models with the GEneral NEural SImulation
%  System.}  New York: TELOS/Springer--Verlag.

%[3] Hasselmo, M.E., Schnell, E.\ \& Barkai, E.\ (1995) Dynamics of
%learning and recall at excitatory recurrent synapses and cholinergic
%modulation in rat hippocampal region CA3. {\it Journal of
%  Neuroscience} {\bf 15}(7):5249-5262.

\end{document}